\pdfoutput=1

\documentclass[11pt]{article}

\usepackage[]{acl}
\usepackage{times}
\usepackage{latexsym}
\usepackage{graphicx}
\usepackage[T1]{fontenc}

\usepackage[utf8]{inputenc}

\usepackage{microtype}

\usepackage{xcolor}
\usepackage{cleveref}

\title{Automated Evaluation for Student Argumentative Writing: A Survey}

\author{Xinyu Wang \\
  Riiid Labs \\
  \texttt{xinyuwang699@gmail.com} \\\And
  Yohan Lee \\
  Riiid Labs  \\
  \texttt{yohan.lee@riiid.co} \\\And
  Juneyoung Park \\
  Riiid AI Research \\
  \texttt{	
juneyoung.park@riiid.co} \\}

\begin{document}
\maketitle
\begin{abstract}
\label{abstract}
This paper surveys and organizes research works in an under-studied area, which we call automated evaluation for student argumentative writing. Unlike traditional automated writing evaluation that focuses on holistic essay scoring, this field is more specific: it focuses on evaluating argumentative essays and offers specific feedback, including argumentation structures, argument strength trait score, etc. The focused and detailed evaluation is useful for helping students acquire important argumentation skill. In this paper we organize existing works around tasks, data and methods. We further experiment with BERT on representative datasets, aiming to provide up-to-date baselines for this field.

\end{abstract}

\section{Introduction}
\label{intro}
Automated writing evaluation uses computer programs to give evaluative feedback to a piece of written work, which is often used in educational settings~\citep{parra2019automated}. 
Most of the work on automated writing evaluation focus on automated essay scoring (AES), which evaluates an essay's quality by assigning scores \cite{shermis2013handbook}. It is a long-standing research problem, with the first system proposed in 1966 by \citet{page1966imminence}. Since then this area has been active and attracts a lot of research efforts.

A large portion of AES systems are developed for \textit{holistic} scoring~\cite{ke2019automated}, which outputs a single score to represent essay quality. 
This kind of system is useful for summative assessment and can greatly reduce manual grading efforts. 
However, holistic scoring is not sufficient for providing instructional feedback to student learning because a low holistic score does not provide enough information to the students about how to improve. 
To address this issue, many research works tried to build \textit{trait-specific} scoring systems. These systems concern scoring particular quality dimensions of an essay, such as grammar \cite{burstein2004automated}, word choice \cite{mathias-bhattacharyya-2020-neural}, coherence \cite{somasundaran2014lexical}, etc. 
Holistic scoring systems, along with trait specific scoring systems, have already been deployed to commercial settings successfully (\citealt{burstein2004automated}, \citealt{rudner2006evaluation}). 

However, most of the systems do not distinguish between \textit{essay types} (\textit{e.g.}~argumentative or narrative essay). It makes sense when the system is trying to evaluate type-agnostic dimensions such as word choice of an essay, but the use of such systems is limited when an user wants to know more about the in-depth traits specific to an essay type, e.g. whether claims and counterclaims are developed fairly in an argumentative essay. The evaluation of type specific traits is important because these traits reflect important type-specific skills --- for example, whether claims and counterclaims are developed thoroughly is indicative of critical thinking skills, which can be hard to gauge through narrative essay writing. 

In this paper we attempt to organize the research works regarding automated evaluation specifically for student argumentative writing, which requires the students to evaluate controversial claims, collect and judge evidence and establish a position. The contributions of this paper are as follows. First, we categorized and organized current research body regarding tasks and datasets (\cref{tasksdata}) and methods (\cref{methods}). Additionally, we experimented with BERT \cite{devlin-etal-2019-bert} models and provided up-to-date baselines to the community (\cref{exp}). Finally, we suggested directions (\cref{future}) based on missing elements in current research body which we hope to be filled in the future. 

\begin{figure*}[t]
\centering
\includegraphics[width=15cm]{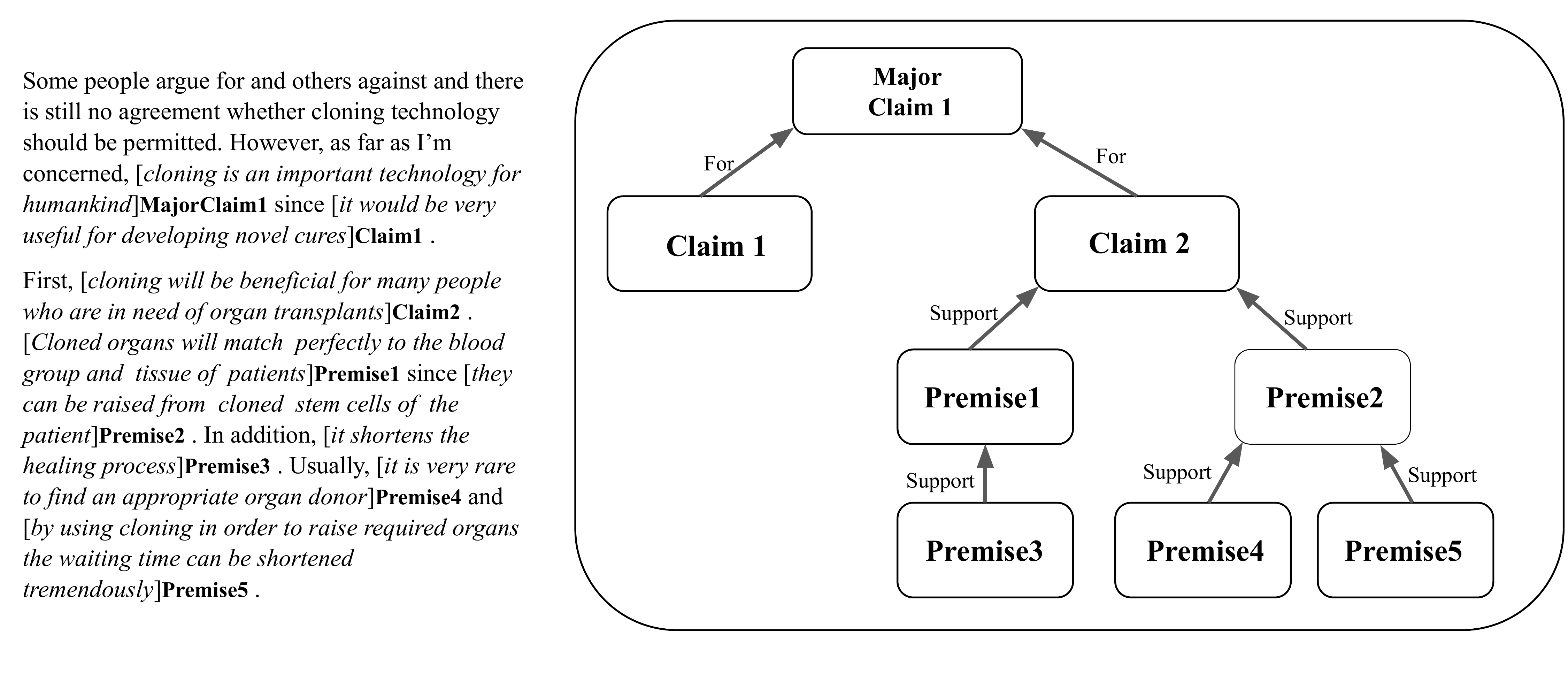}
\caption{An illustration of annotation scheme for S\&G2014 and S\&G2017a. This figure is adapted from \citet{stab2017-parsing} }
\label{exampleAnno}
\end{figure*}

\begin{table}[]
\centering
\begin{tabular}{l|l}
\hline
\textbf{Type}           & \textbf{Example Tasks}                                                                               \\ \hline
AM    & Argument Component Identification             \\
                    & Argument Component Classification           \\
                    & Argument Relation Identification            \\
                    & Relation Labelling                     \\
                    \hline
CD & Opposing Argument Detection       \\
                    & Valid Critique Detection                  \\
                    & Thesis Detection \\
                    \hline
QA  & Sufficiency Recognition                    \\
                    & Argument Strength Scoring                     
\end{tabular}
\caption{Example sub-tasks under the three main problems: argument mining (AM), component detection (CD) and quality assessment (QA)}
\label{taskDesc}
\end{table}




\section{Tasks and Datasets}
\label{tasksdata}
In this section, we introduce the tasks commonly studied in automated writing evaluation for argumentative writing, together with their benchmark datasets.
\begin{table*}[t]
\centering
\begin{tabular}{llll}
\hline
\textbf{Data} & \textbf{No. of Essays} & \textbf{Label Structure} & \textbf{Task Type}\\
\hline
\citealt{stab-gurevych-2014-annotating} & 90   & Tree & Argument mining \\ 
\citealt{stab2017-parsing} & 402   & Tree & Argument mining \\ 
\citealt{stab2016} & 402 & Binary value & Component Detection \\
\citealt{carlile-etal-2018-give} & 102  & Values on top of tree nodes & Quality Assessment \\ 
\citealt{stab2017-sufficiency} & 402  & Binary value & Quality Assessment \\ 
\citealt{persing-ng-2015-modeling} & 1000  & Score & Quality Assessment \\
\hline
\end{tabular}
\caption{\label{citation-guide}
Comparison of several popular public datasets
}
\end{table*}

\subsection{Argument Mining}
\label{argmining}
Argument mining (AM) aims to identify and parse the argumentation structure of a piece of text. \cite{lawrence2020argument}
For argumentative essays, argumentation structures may have variations but can typically be represented as trees, as illustrated in Figure \ref{exampleAnno}.
The root of the tree is a major claim, which expresses the author's main point on the topic.
Children of the major claims are claims, which are controversial statements that either argue for or against their parents.
Claims could have children nodes, called premises (\textit{e.g.}~$Premise1$ and $Premise3$ on Figure \ref{exampleAnno}), which support or attack the corresponding claims. 
Further, a premise (\textit{e.g.}~$Premise3$) can also be supported or attacked by other premises (\textit{e.g.}~$Premise4$ and $Premise5$), enriching the logical flow of the essay.

In order to generate an argumentation structure from text, argument mining models typically decompose the process into four sub-tasks: 
(1) argument component identification (ACI), which aims to identify the span of argument components (\textit{e.g.}~text with squared brackets on Figure \ref{exampleAnno}); (2) argument component classification (ACC), which classifies argument components into corresponding types (\textit{e.g.}~where a span is a {\tt Claim} or {\tt Premise}); 
3) argument relation identification (ARI), which aims at linking related argument components on the argument structure; 4) relation labelling (RL), which focuses on identifying relation type (\textit{e.g.}~\textit{for} and \textit{support}) between linked components. 

\paragraph{Datasets} 
\citealp{stab-gurevych-2014-annotating} (S\&G2014) and \citealp{stab2017-parsing} (S\&G2017a) annotated the most widely adopted datasets regarding this direction. 
S\&G2014 annotated 90 persuasive essays from \textit{essayforum}, a site providing feedback for students who wish to improve their writing.
S\&G2017a adopted similar annotation scheme and annotated additional 402 essays.  


Besides, \citet{putra2021annotating} annotated 434 essays written by English to Speakers of Other Languages (ESOL) college learners as quasi-trees. They annotated each sentence as either argumentative component (ACs) or non-argumentative component (non-AC). Annotators then identified relations between ACs. Different from S\&G2014, their relation labels include directed labels (support, attack or detailing) as well as an undirected label(restatement). They also reordered the sentences to make the essays better-structured. \citet{wambsganss2020corpus} annotated 1000 student peer-reviews written in German, indicating whether a text span is claim or premise or not, and the relations between these argumentative components. 

\citet{alhindi2021} annotated on a token level using BIO tagging schema and this dataset can be used for ACI and ACC. Specifically, each token belongs to one of the five classes: the begin token of a claim; a continuous token of a claim; the begin token of a premise; a continuous token of a premise; non-argumentative token.  

\subsection{Component Detection}

Besides parsing the complex argumentative structures of essays as in argument mining, a large portion of work in automated writing evaluation also considers the simpler task of detecting whether an essay or a text span in an essay contains a specific component. 
For example, \citet{stab2016} assigned binary labels to essays depending on whether an essay has discussed the opposing arguments to author's own standpoint.
\citet{falakmasir} labelled essays as whether or not containing a thesis and conclusion statement.
\cite{klebanov2017} annotated essays written by college students for criticizing a piece of argument, deciding if each sentence contains a good critique or not. \cite{ghosh2020-exploratory} annotated on whether a sentence contains a valid critique as well, but the essays they used are written by middle school students. 

Finally, we remark that component detection is different from the task of argument component classification (ACC) discussed previously in Section \ref{argmining}.
Specifically, component detection does not require the text span to match the component of interest. For example, \citet{falakmasir} cares about whether an essay contains a thesis and conclusion statement. In their case, the length of text span is much longer than the component of interest. 


\subsection{Quality Assessment}

Quality assessment concerns evaluating the quality of an argument. There are many sub-tasks under this category due to the diversity of argumentation quality theories.

\citet{persing-ng-2015-modeling} annotated 1000 essays over 10 prompts from International Corpus of Learner English (ICLE) dataset \cite{granger2009international} on argument strength. They defined argument strength as how well an essay makes an argument for its thesis and convinces its readers. \citet{horbach2017fine} collected 2020 German essays written by prospective university students. Their annotators evaluated the text regarding 41 aspects including quality of argumentation.  In addition, \citet{stab2017-sufficiency} (S\&G2017b) adopted Relevance-Acceptability-Sufficiency criteria \citep{johnson2006logical} and asked annotators to decide whether a piece of argument is sufficiently supported or not. 


Besides, \citet{carlile-etal-2018-give} annotated detailed persuasiveness and attributes values based on argument trees from S\&G2014 and S\&G2017a. They defined an argument as a node in the argument tree along with all its children. For each argument, they assigned an overall persuasive score, common attribute values and type specific attribute values. For example, for arguments using a major claim as the root node, annotations include persuasive score, \textit{eloquence}, \textit{specificity}, \textit{evidence} (common attributes) and persuasive strategies (type-specific attribute).

                    

\section{Methods}
\label{methods}
All the methods to our knowledge are learning based and they can be divided into supervised learning and unsupervised learning. Most of the work use supervised learning, which can be further divided into feature-based approaches and neural approaches. Next, we introduce these approaches in more details.

\subsection{Feature-based method}
For feature-based approaches, off-the-shelf algorithms are typically used for model training on hand-crafted input features. For example, support vector machine (SVM), logistic regression and random forest are typically used for tasks that can be framed as classification (\citealp{stab2014-identifying}, \citealp{stab2017-sufficiency}, \citealp{stab2017-parsing}, \citealp{persing-ng-2016-end}, \citealp{klebanov2017}, \citealp{wan2021}). Linear regression and support vector machine regression are often used for task that can be framed as regression (\citealp{persing-ng-2015-modeling}, \citealp{wachsmuth-etal-2016-using}, \citealp{persing-ng-2015-modeling}). 

Next, we introduce more details of common features used in these methods. 

\textbf{Lexical features}
aim to capture word-level information and common lexical features include n-gram and frequent words. They have been shown effective (\citealp{stab2014-identifying}, \citealp{klebanov2017}, \citealp{stab2017-sufficiency}). However, they do not perform as well when used in a \textit{cross-prompt} setting \cite{klebanov2017} where prompts of testing essays are not seen during training. This is intuitive as the actual wording of essays of different prompts would differ significantly.

\textbf{Syntactic features}
usually rely on parse trees. Common syntactic features include number of sub-clauses in a parse tree, Boolean indicator of production rules, part-of-speech tags, etc. Additionally, basic information such as tense of verbs, presence of modal verbs can also serve as syntactic features. \cite{stab2017-parsing} showed that syntactic features are useful for identifying the beginning of an argument component and \cite{stab2017-sufficiency} suggested that syntactic features are effective for recognizing insufficiently supported arguments. 

\textbf{Structural features}
generally describe the position and frequency of a piece of text. For example, they include position of a token, punctuation and an argument component. They also include statistics such as number of tokens in an argument component. \cite{stab2017-parsing} reported effectiveness for these features on both ACI and ACC tasks. 

\textbf{Embedding features}
are based on word vectors that represent words in a continuous space and are supposed to capture more information than simple n-grams. \citet{stab2017-parsing} summed the word2vec \cite{mikolov2013distributed} vectors for each token to represent a component. \citet{putra-etal-2021-parsing} used BERT \cite{devlin-etal-2019-bert} to extract token embeddings in their work.

\textbf{Discourse features}
captures how sentences or clauses are connected together. One kind of discourse features depend on discourse markers directly. The markers, such as "therefore", suggest the relationship between current text span and its adjacent text span. Another kind of discourse features use the output of discourse parsers. For example, \citet{klebanov2017} parsed sentences into corresponding discourse roles, and then used these discourse roles as features. \citet{stab2017-parsing} reported usefulness of discourse features on classifying argument components, indicating a correlation between general discourse relation and argument component type. \citet{klebanov2017} found that discourse features remain useful in cross-prompt settings, which is valuable as it's not always possible to collect a lot of data for a single prompt.

\subsection{Neural method}
As for \textbf{neural} approaches, neural architectures such as long short-term memory (LSTM) networks and convolutional neural networks (CNN) are commonly adopted (\citealp{Eger:2017:ACL}, \citealp{alhindi2021}, \citealp{putra-etal-2021-parsing}, \citealp{discoshuffle}, \citealp{xue2020},\citealp{stab2017-sufficiency}). Besides, Transformer \cite{vaswani2017attention} based architectures such as BERT \cite{devlin-etal-2019-bert} are adopted recently (\citealp{ye-teufel-2021-end}, \citealp{putra-etal-2021-parsing}, \citealp{ghosh2020-exploratory}, \citealp{alhindi2021} \citealp{wang-etal-2020-argumentation}). We will describe more details of that later on. 

\textbf{Use of pretrained models}
The use of pretrained language models has been popular among natural language processing community. This is because state-of-the-art models are pretrained on massive text corpus, allowing information learned from a huge amount of text to be used for downstream tasks. 

There are two ways to use pretrained models. First, it can be used as feature extractor. For example, \citet{putra-etal-2021-parsing} has used BERT, a bi-directional Transformer based architecture, to extract input embeddings, and then pass the contextualized embeddings to downstream networks. Second, the pretrained language models can be further fine-tuned for the task of interest. \citet{ye-teufel-2021-end}, \citet{wang-etal-2020-argumentation}, \citet{alhindi2021} and \citet{ghosh2020-exploratory} used this approach. In addition, other than fine-tuning on the task data directly, \citet{alhindi2021} and \citet{ghosh2020-exploratory} experimented with continued pre-training with a large unlabelled domain relevant corpus first, and then fine-tuning with task data.


\subsection{Address multiple sub-tasks}
\label{subtaskmethod}
As mentioned, argument mining mostly concerns four sub-tasks and they are often addressed together. The naive way to solve them simultaneously is to model each sub-task separately in a pipeline fashion. This introduces at least two issues: first, it does not enforce any constraints between different tasks; second, the errors made early on in the pipeline could propagate. 

One way to address these is to use \textit{integer linear programming} (ILP) \cite{roth2004linear} for joint inference. \citet{stab2017-parsing} and \citet{persing-ng-2016-end} adopted this approach and \citet{persing-ng-2016-end} further proposed an ILP objective that directly optimizes F score.

\begin{figure}
\centering
\includegraphics[width=7.5cm]{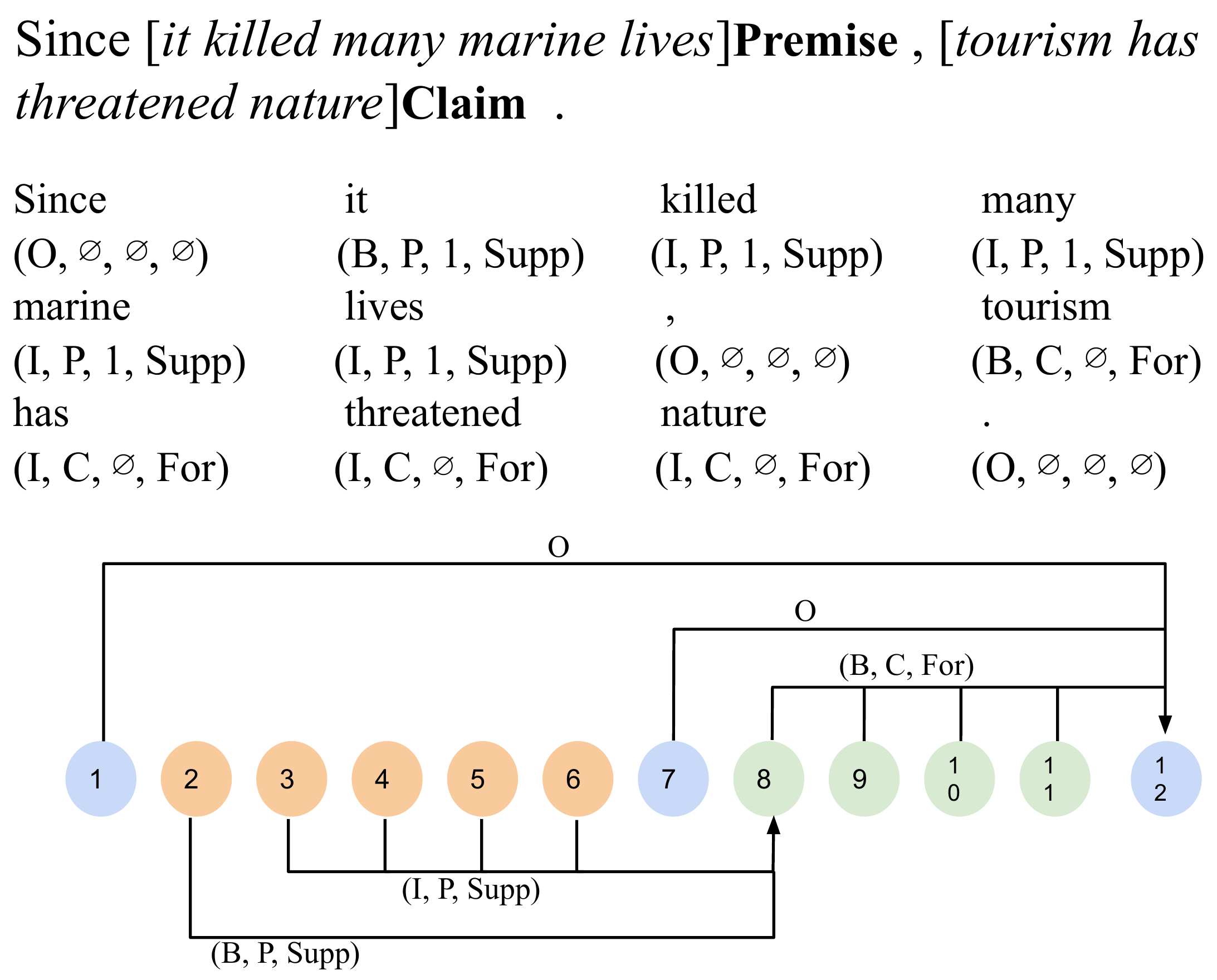}
\caption{Illustration of the formulations described in \ref{subtaskmethod}. On the top is an annotated sentence. The second chunk represents sequence tagging formulation and the bottom is dependency parsing formulation. In the figure, B means the begin of a component, I means the continue of a component, C stands for Claim, P stands for Premise, O stands for non-argumentative, Supp stands for Support and $\emptyset$ means not filled. Figure is adapted from \citet{Eger:2017:ACL}}
\label{amframing}
\end{figure}

Another way to address these issues is to build a joint model that can model all these tasks at the same time. \citet{Eger:2017:ACL} proposed two different frameworks for joint modelling the full argument mining task and has been influential. Figure \ref{amframing} showed an illustration of these two formulations. They first framed the task as \textit{sequence tagging}. For example, they used S\&G2017a data for experiments and in this case, each token's label space would include 1) whether this token is a begin or continue of an argument component or it is non-argumentative); 2) the type of the component to which the token belongs; 3) distance between the corresponding component and the component it relates to; 4) the relation type between the two related components. This way they can use off-the-shelf taggers to solve all four sub-tasks at once. They also framed the task as \textit{dependency parsing}. When framed as dependency parsing, the text is represented as directed trees where each token has a labelled head so that argument component relation information can be encoded. They further labelled these edges with tokens' component types and relation types so that argument trees can be converted to quasi-dependency trees. Finally, they adapted a \textit{joint neural model} designed for entity detection and relation extraction \cite{miwa-bansal-2016-end}. In this case, they modelled argument components as entities and argument relations as semantic relations.

\subsection{Unsupervised learning}
Aside from the supervised learning work described above, \citet{persing-ng-2020-unsupervised} proposed an unsupervised method for argument mining. The key for their work is to use heuristics for bootstrapping a small set of labels and then train the model in a self-training fashion.

\section{Experiments}
\label{exp}
The use of Transformer-based models has been dominating in other natural language processing tasks but is still in its infant stage in this field. Therefore, we experiment with vanilla BERT models on three representative datasets, hoping to facilitate research to that end. The three datasets cover argumentation structure parsing and argument quality assessment. Specifically, we used S\&G2017a dataset, which is the benchmark for parsing essay argumentation structure. Besides, we used S\&G2017b dataset, which assess argument quality from a logic aspect \cite{wachsmuth-etal-2017-computational} by annotating whether a piece of argument is sufficiently supported. We also used P\&N2015 dataset which assess argument quality from a rhetoric aspect \cite{wachsmuth-etal-2017-computational} by assigning an overall argument strength score to each essay. Statistics for these datasets can be found in Table \ref{SG2017a}, \ref{SG2017b}, \ref{PN2015} and \ref{scoredistr}.
\subsection{Implementation details}

\begin{table}[]
\centering
\begin{tabular}{l|ll}
\hline
          & All     & Per Essay \\ \hline
Token     & 147,271 & 366.3   \\
Sentence  & 7,116    & 17.7    \\
Paragraph & 1,833    & 4.6    \\
Essay & 402 & 1
\end{tabular}
\caption{Data statistics for the whole S\&G2017a dataset.}
\label{SG2017a}
\end{table}
\textbf{S\&G2017a} For S\&G2017a dataset, we followed \citet{Eger:2017:ACL} and used 286 essays for training, 36 essays for validation and 80 essays for testing on paragraph level. For model building, we used Huggingface \cite{wolf-etal-2020-transformers} library's \textit{cased base BERT} model. We added one shared dropout layer and three linear heads for predicting labels mentioned in \ref{subtaskmethod}. The loss is computed by summing up the cross entropy loss on each head. Note that BERT's tokenizer can tokenize each word into multiple sub-tokens but we want to predict only one set of tags for each word. To address this, we only use the first sub-token for training. For model training, we used AdamW optimizer \cite{loshchilov2017decoupled}, learning rate 3e-5 and Cosine Annealing Warm Restart scheduler \cite{loshchilov2016sgdr} implemented by PyTorch Lightning library \cite{falcon2019pytorch}. We also monitored validation loss for early stopping. We only tuned dropout rate and patience used for early stopping minimally and we set patience to 5 and dropout rate to 0.5.

\begin{table}[]
\centering
\begin{tabular}{l|ll}
\hline
             & All    & Per Argument \\ \hline
Token        & 97,370 & 94.6         \\
Sentence     & 4,593  & 4.5          \\
Argument     & 1,029  & 1          \\ \hline
             & Number & Percentage   \\ \hline
Sufficient   & 681    & 66.2\%      \\
Insufficient & 348    & 33.8\%     
\end{tabular}
\caption{Data statistics for S\&G2017b dataset. It includes the size of the corpus and class distributions.}
\label{SG2017b}
\end{table}
\textbf{S\&G2017b} For S\&G2017b dataset, we used 823 arguments for training, 103 arguments for validation and 103 arguments for testing. We used the same BERT model and added a dropout layer and a linear layer. We used binary cross entropy loss as objective function. For model hyperparameters, we set patience to 5, dropout rate to 0.3 and kept everything else the same as S\&G2017a.

\begin{table}[]
\centering
\begin{tabular}{l|ll}
\hline
          & All     & Per Essay \\ \hline
Token     & 649,549 & 649.5     \\
Sentence  & 31,589  & 31.6      \\
Paragraph & 7,537   & 7.5       \\
Essay     & 1,000   & 1        
\end{tabular}
\caption{Data statistics for P\&N2015 dataset.}
\label{PN2015}
\end{table}

\begin{table}[]
\centering
\setlength{\tabcolsep}{1.2mm}{
\begin{tabular}{l|l|l|l|l|l|l|l}
\hline
Score     & 1.0 & 1.5 & 2.0   & 2.5 & 3.0   & 3.5 & 4.5  \\ \hline
Essays & 2 & 21  & 116 & 342 & 372 & 132 & 15
\end{tabular}}
\caption{Score distribution for P\&N2015 dataset.}
\label{scoredistr}
\end{table}
\textbf{P\&N2015} For P\&N2015 dataset, we used 800 essays for training, 100 essays for validation and 100 essays for testing. Additionally, we ran cross-prompt experiments by randomly selecting 247 essays from prompt 1 for training, 31 essays for validation, and rest of prompt 1 essays as well as essays from other prompts for testing. We also used one dropout layer and one linear layer on top of the base BERT model. For objective function, we used mean square loss. Besides, we set patience to 5 and dropout rate to 0.1 and kept everything else the same as S\&G2017a.

\subsection{Metrics and comparison methods}
\textbf{S\&G2017a} Similar to most work, we followed \citet{persing-ng-2016-end} for evaluation. They defined F1 score as F1 = $\frac{2TP}{2TP+FP+FN}$, where TP stands for true positive, FP stands for false positive and FN stands for false negative. They also define 'level $\alpha$ matching' \cite{Eger:2017:ACL}: for $\alpha$\% level match, the predicted component span and ground truth share at least $\alpha$\% of their tokens. For comparison methods, we chose the LSMT-ER model from \citet{Eger:2017:ACL}, which is a common baseline. In addition, we compared with the BiPAM model from \citet{ye-teufel-2021-end}, which is a BERT-enhanced biaffine dependency parser \cite{dozat-manning-2018-simpler}.

\textbf{S\&G2017b} We used macro F1 and accuracy scores for evaluation, following \citet{stab2017-sufficiency}. As for comparison methods, we chose the best performing CNN model in their work and human upper bound. 

\textbf{P\&N2015} We used mean absolute error (MAE) and mean square error (MSE) for evaluation. We compared the baseline with a model developed by \citet{wachsmuth-etal-2016-using}, which is the best performing model to our knowledge.

\subsection{Results and discussion}
\label{discuss}
\textbf{S\&G2017a} The experiment results for this argument mining task are shown in Table \ref{AMres}. The vanilla BERT did not outperform state-of-the-art methods but are close. The BiPAM model, which is a BERT-enhanced dependency parser, performs the best. These demonstrate the power of Transformer-based models for the argument mining task. Additionally, LSTM-ER has 1) designed separate module for handling components and relations; 2) encoded explicit syntax information through syntactic parser. BiPAM has carefully designed the argument relation representation so that it can benefit from state-of-the-art dependency parsers. These considerations can be used for further unlocking the potential of Transformer-based models. 

\begin{table}[]
\centering
\begin{tabular}{l|rr|rr}
\hline
                                    & \multicolumn{2}{c|}{C-F1}        & \multicolumn{2}{c}{R-F1}        \\ \cline{2-5} 
                                    & 100\% & 50\%                     & 100\% & 50\%                    \\ \hline
LSTM-ER                             & 70.8     & \textbf{77.2}                        & 45.5     & \textbf{50.1}                       \\
BiPAM                               & \textbf{72.9}     & \multicolumn{1}{l|}{N/A} & \textbf{45.9}     & \multicolumn{1}{l}{N/A} \\
BERT+linear & 69.3     & 76.7                        & 43.7     & 47.6                      
\end{tabular}
\caption{Performance of LSTM-ER, BiPAM and vanilla BERT on S\&G2017a dataset. The models are all trained on paragraph level and we report both 100\% level match and 50\% level match results. C-F1 stands for argument component F1 and R-F1 stands for argument relation F1. Best scores are in bold. }
\label{AMres}
\end{table}

\textbf{S\&G2017b} The experiment results for sufficiency recognition are shown in Table \ref{sufficiencyRes}. We can see that the vanilla BERT model outperforms the previous best performing CNN model by a large margin. This is expected as the BERT model has been pretrained on a huge amount of text. It is surprising that the vanilla BERT model already achieved near-human performance. Therefore, we foresee that Transformer-based model can equal or even surpass human upper bound on sufficiency recognition in the near future.  

\begin{table}[]
\centering
\begin{tabular}{l|ll}
\hline
                  & Accuracy    & Macro F1    \\ \hline
Human & \textbf{0.911±.022}  & \textbf{0.887±.026}  \\
CNN               & 0.843±.025  & 0.827±.027  \\
BERT+linear       & 0.882±.018 & 0.869±.012
\end{tabular}
\caption{Performance of CNN, vanillaBERT and human upper bound on S\&G2017b. Best scores are in bold.}
\label{sufficiencyRes}
\end{table}


\textbf{P\&N2015}
The experiment results for argument strength scoring are shown in Table \ref{ARSRes}. On this dataset, the vanilla BERT model did not outperform previous methods. There are two possible reasons for explaining it. First, the average token in an single essay exceed the maximum length(512) supported by vanilla BERT model. This results in part of the essay being truncated. From the data statistics, we can know that at least 25\% of the essay is being truncated, resulting in large information loss. Second, both \citet{persing-ng-2015-modeling} and \citet{wachsmuth-etal-2016-using} encoded argument structure information explicitly by crafting a list of argument structure related features. \citet{mim-etal-2019-unsupervised} incorporated paragraph's argument function information by pretraining on large-scale essays with shuffled paragraphs. At the same time, BERT model was not pretrained on relevant task and might fail at capturing the overall argument structure of an essay.  

\begin{table}[h]
\centering
\setlength{\tabcolsep}{4.8mm}{
\begin{tabular}{l|ll}
\hline
            & MAE         & MSE         \\ \hline
Persing2015   & 0.392   & 0.244 \\
Wachsmuth2016   & \textbf{0.378} & \textbf{0.226} \\
Mim2019   & N/A & 0.231 \\
BERT+linear & 0.394 & 0.250
\end{tabular}}
\caption{Performance of best models from \citet{persing-ng-2015-modeling}, \citet{wachsmuth-etal-2016-using} and \citet{mim-etal-2019-unsupervised}, and vanillaBERT on P\&N2015. }
\label{ARSRes}
\end{table}

\begin{table*}[]
\centering
\begin{tabular}{l|l|l|l|l|l|l|l|l|l|l}
\hline
Test Prompt    & 1     & 2     & 3     & 4     & 5     & 6     & 7     & 8     & 9     & 10    \\ \hline
MAE & 0.421 & 0.399 & 0.485 & 0.443 & 0.471 & 0.389 & 0.371 & \textbf{0.369} & 0.398 & 0.439 \\
MSE & 0.237 & 0.268 & 0.349 & 0.286 & 0.403 & 0.225 & \textbf{0.217} & 0.246 & 0.251 & 0.283
\end{tabular}
\caption{Results of cross-prompt experiments on P\&N2015 dataset}
\label{crossprompt}
\end{table*}

\begin{figure}
\includegraphics[width=7.5cm]{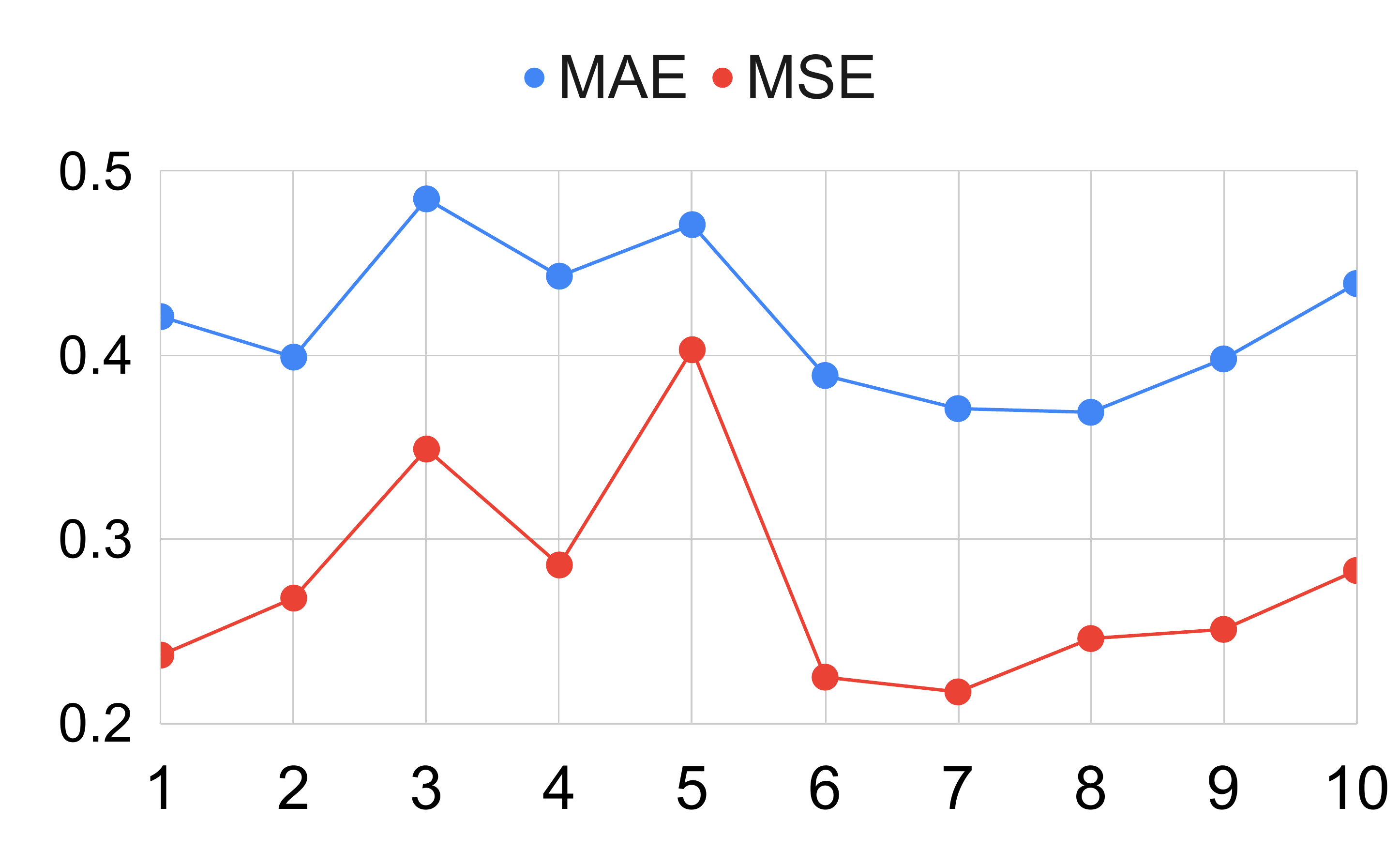}
\caption{Line chart based on Table \ref{crossprompt}}
\label{crosspromptfig}
\end{figure}

To further gauge the generalization ability of Transformer-based models, we ran the vanilla BERT in a cross-prompt setting and the results are shown in \ref{crossprompt}. Being able to generalize across prompt is valuable in practice because it is expensive to collect data for each new prompt. Recall that the model is trained and validated on essays of prompt 1. From \ref{crosspromptfig}, we can see that the MSE and MAE remains similar across prompts except for prompt 3 and prompt 5. We took a look at these two prompts and found that they are much more abstract and provide less concrete context than the other prompts. \footnote{Due to licensing of the ICLE dataset, we cannot provide the detailed information here.} We hypothesize it to be the performance drop on prompt 3 and prompt 5 and thus believe that BERT can generalize reasonably well.

Overall, the vanilla BERT models perform comparably to previous methods across datasets. We believe this can be a good baseline model for future research in the community.


\section{Conclusion and Future Directions}
\label{future}
In this paper we have organize existing works around tasks, data and methods regarding automated evaluation for student argumentative writing. We provided a baseline to the community by experimenting with BERT models on benchmark datasets. Now the question would be: what is important for the future? Towards this end, we identified several directions we deem interesting. 

First of all, there is a lack of emphasis on language diversity in current datasets. This concerns two aspect: language of the essays and language backgrounds of the authors. For language of the essays, most datasets are annotated essays written in English. \citet{horbach2017fine} and \citet{wambsganss2020corpus} annotated essays written in German and these two datasets are the only ones not written in English to our knowledge. Intuitively speaking, characteristics such as overall argumentation structure would vary across essays written in different languages. Besides, annotators for essays written in different languages may have different criteria when assessing the rhetoric quality of an essay. As for authors' language backgrounds, current datasets either do not take into account authors' language levels or use essays written by proficient writers (e.g. college-level writing). \footnote{Although P\&N2015 dataset annotated English learner's texts, most of the authors' native languages belong to Indo-European languages and have received at least 6 or more years' English education} \citet{putra2021annotating} and \citet{alhindi2021} are the only accessible datasets that target authors whose interested language skills are not proficient yet, to our knowledge. Specifically, \citet{putra2021annotating} annotated essays written by English learners from various Asian countries. They discarded already well-written essays and only reserved the ones with intermediate quality. \citet{alhindi2021} annotated argumentative essays by middle school students and found these essays less structured and thus more challenging. We believe collecting more diverse datasets would be valuable because it will not only expand the impact of argumentative writing support systems, but also pose more challenging research problems. 

Second, adoption of Transformer-based models is still in the infant stage despite some recent works have used Transformer-based architectures. As described in \ref{exp}, we have built the simplest form of BERT models and demonstrated that they have comparable performance to previous state-of-the-art methods. In addition, \citet{ghosh2020-exploratory} and \citet{alhindi2021} showed that performance of BERT can be greatly improved by continued pre-training on unlabelled essay datasets or architectural design that takes account of data characteristics. Therefore, we believe there is still huge potential for Transformer-based models.

Third, there is no research regarding generalization in this field. We are aware of relevant research works in the AES area (\citealt{jin2018tdnn}, \citealt{cao2020domain}, \citealt{ridley2021automated}) but not for systems that care about argument-related attributes. Not being able to generalize across prompts remains a major bottleneck in AES \cite{woods2017formative} and we believe this can also be a main obstacle for deploying automated evaluation systems specifically for argumentative essays. This is because it is costly to collect data for every prompt. At the same time, learners usually need to practice writing over a large amount of prompts and get feedback before seeing significant improvement. This echoes our promotion for Transformer-based models as we have shown in \ref{discuss} a vanilla BERT model can generalize well across prompts within one dataset. Besides, Transformer-based models have shown to be effective on unseen domains in other NLP tasks (\citealt{houlsby2019parameter}, \citealt{han2019unsupervised}). Overall, we believe it is critical to build generalizeable systems and hope to see more research addressing this issue in the future.


\bibliographystyle{acl_natbib}
\bibliography{anthology, custom}

\appendix



\end{document}